Article

# Energy-Efficient Inference on the Edge Exploiting TinyML Capabilities for UAVs

Wamiq Raza *,†, Anas Osman † , Francesco Ferrini and Francesco De Natale

Department of Information Engineering and Computer Science—DISI, University of Trento, 38122 Trento, Italy; anas.osman@studenti.unitn.it (A.O.); francesco.ferrini@studenti.unitn.it (F.F.); francesco.denatale@unitn.it (F.D.N.)
* Correspondence: wamiq.raza@studenti.unitn.it
† These authors contributed equally to this work.

**Abstract:** In recent years, the proliferation of unmanned aerial vehicles (UAVs) has increased dramatically. UAVs can accomplish complex or dangerous tasks in a reliable and cost-effective way but are still limited by power consumption problems, which pose serious constraints on the flight duration and completion of energy-demanding tasks. The possibility of providing UAVs with advanced decision-making capabilities in an energy-effective way would be extremely beneficial. In this paper, we propose a practical solution to this problem that exploits deep learning on the edge. The developed system integrates an OpenMV microcontroller into a DJI Tello Micro Aerial Vehicle (MAV). The microcontroller hosts a set of machine learning-enabled inference tools that cooperate to control the navigation of the drone and complete a given mission objective. The goal of this approach is to leverage the new opportunistic features of TinyML through OpenMV including offline inference, low latency, energy efficiency, and data security. The approach is successfully validated on a practical application consisting of the onboard detection of people wearing protection masks in a crowded environment.

## 1. Introduction

Drones, in the form of both Remotely Piloted Aerial Systems (RPAS) and unmanned aerial vehicles (UAV), are increasingly being used to revolutionize many existing applications. The Internet of Things (IoT) is becoming more ubiquitous every day, thanks to the widespread adoption and integration of mobile robots into IoT ecosystems. As the world becomes more dependent on technology, there is a growing need for autonomous systems that support the activities and mitigate the risks for human operators [1]. In this context, UAVs are becoming increasingly popular in a range of civil and military applications such as smart agriculture [2], defense [3], construction site monitoring [4], and environmental monitoring [5].

These aerial vehicles are subject to numerous limitations such as safety, energy, weight, and space requirements. Electrically powered UAVs, which represent the majority of micro aerial vehicles, show a severe limitation in the duration of batteries, which are necessarily small due to design constraints. This problem affects both the flight duration and the capability of performing fast maneuvers (e.g., to avoid obstacles) due to the slow power response of the battery. Therefore, despite their unique capabilities and virtually unlimited opportunities, the practical application of UAVs still suffers from significant restrictions [6].

Recent advances in embedded systems through IoT devices could open new and interesting possibilities in this domain. Edge computing brings new insights into existing IoT environments by solving many critical challenges. Deep learning (DL) at the edge presents significant advantages with respect to its distributed counterpart: it allows the performance of complex inference tasks without the need to connect to the cloud, resulting in a significant latency reduction; it ensures data protection by eliminating the vulnerability connected to the constant exchange of data; and it reduces energy consumption by avoiding the transmission of data between the device and the server [7].

Another recent trend refers to the possibility of shifting the ML inference peripherally by exploiting new classes of microcontrollers, thus generating the notion of Tiny Machine Learning (TinyML) [8]. TinyML aims to bring ML inference into devices characterized by a very low power consumption. This enables intelligent functions on tiny and portable devices with a power consumption of less than 1 mW. As TinyML targets microcontroller unit (MCU) class devices, the trained and developed models must conform to the hardware and software constraints of MCUs. Therefore, many TinyML frameworks have been developed to fully exploit MCU resources and ensure the optimization of the model converted from the original model, e.g., the TensorFlow Lite for Microcontrollers (TFLM) framework. TFLM is one of the most widely used TinyML frameworks, which is compatible with the well-known ML libraries TensorFlow [9] and Keras [10].

Building upon the above technological trends, the integration of state-of-the-art ultra-low power embedded devices into UAVs could provide energy-aware solutions to embed an increasing amount of autonomy and intelligence into the drone, thus paving the way for many novel and versatile applications. Such devices can run on a coin-sized battery, are capable of processing large amounts of data in real-time, and, most importantly, can perform inferences at the edge without requiring an external connection to the cloud or to a remote processing unit [11], thus avoiding the need for energy-eager data transmission protocols.

In this paper, we propose a novel approach to endow drones with deep learning capabilities without compromising flight time or imposing further constraints. We present the integration of an OpenMV Cam H7 MCU, which supports the TFLM, into a DJI Tello drone to enable the development and deployment of TinyML applications on a microcontroller on board the drone. The inference engine implemented on the microcontroller takes care of both the navigation and recognition tasks in an integrated way. The drone flight is controlled by the onboard intelligence and is stabilized through a PID controller to reduce turbulence and obtain steady images. At the same time, the images captured by the camera are processed on board in real-time to accomplish a given high-level task. The main contribution of this work is to demonstrate through a practical implementation the feasibility of a completely autonomous smart drone capable of performing complex missions (in the sample application, targeting people in a crowded environment and detecting if they are wearing a protection mask or not) in an energy-effective way and without the need for connecting to the ground during the flight.

The rest of the paper is organized as follows: in Section 2, we discuss and compare the energy-efficient techniques for UAVs; in Section 3, we present the proposed system; in Section 4, we discuss the system validation by describing an experimental setup and the relevant simulation results; and finally, in Section 5, we draw the conclusions.

## 2. Energy-Efficient Machine Learning Approaches for UAVs

In this section, we briefly review several recent applications of machine learning and artificial intelligence to UAVs, mainly focusing on energy efficiency. In [12], the authors present a long short-term memory (LSTM) inference algorithm to predict future mobile traffic using back-propagation-based neural network training. They proposed the division of the entire coverage area into clusters using a joint k-means and an EM algorithm of a Gaussian mixture model. As their approach required a central intelligence, the optimization was performed at the ground station. Q-learning is another technique used in many UAV applications to find an optimal flight path and recharge method that maximizes the total flight duration of the UAV. As an example, in [13], the UAV received updated information at each step and a UAV base station was used for battery recharging by a wireless power transfer from another flight source. Although the proposed scheme probably required onboard computation, no implementation details were provided because the authors based their analysis on simulations. A meta-gradient inspired ML model over a wireless network was proposed in [14], called hierarchical nested personalized federated learning (HN-PFL). Such a technique optimizes the trade-off between energy consumption and the performance of a machine learning model by configuring the exchange of data among UAVs.

Clustering is a well-known energy-efficient method in which nodes select a cluster head. In [15], a deep learning model is proposed to manage cluster-based UAV networks in an energy-effective way. The model adopts a cluster-level fuzzy logic technique based on a residual network. They compare the achieved results with other competing methods, demonstrating that the T2FL-C guarantees a lower energy consumption; however, an image analysis is performed at the base station where the images are transmitted. Energy-efficient, a fair 3D deployment, and energy replenishment strategies for multiple UAVs are jointly studied in [16]. They take inspiration from deep reinforcement learning (DRL) to design a UAV control policy based on the deep deterministic policy gradient (DDPG), a deep actor critical algorithm. The results are presented based on simulations.

To maximize energy efficiency in terrestrial communications, [17] proposes various key performance indicators (KPIs) sought to satisfy the DRL framework and define an optimal energy efficiency strategy from heterogeneous data. They convert the deep learning problem into deep queue learning to optimize the energy efficiency for airborne users whilst minimizing interference with terrestrial users. Unfortunately, their approach implies a highly complex, non-convex optimization problem and, due to the high mobility, the UAV networks must co-exist with terrestrial networks. The proposed approach is completely centralized as it requires the interconnection of the UAVs to a network of interconnected base stations, which are coordinated by a central control node.

A comparison of the battery-based energy-efficient management solutions is presented in Table 1.

**Table 1.** Comparison between the proposed energy-efficient methods using ML, DL, and RL approaches.

| Research Area | Objective | Outcome | Limitations | Refs. |
|---|---|---|---|---|
| Machine Learning Flying Base Station | The model uses back-propagation learning to predict future traffic | The simulation result shows that the power can be reduced by 24% | Every base station (BS) has the same cellular traffic distribution | [12] |
| Q-Learning at a Recharge Hotspot | An optimized trajectory is proposed to solve the problem. A Markov decision process is used for wireless UAV hotspot services | The results confirm that the two benchmark strategies, random motion and static levitation, outperform SoA | Due to the powerful EM wave requirement, the energy transfer is limited by distance | [13] |
| Distributed ML on UAV Networks For Geo-Distributed Device Clusters | ML personalized local models HN-PFL are proposed where it splits the drone-based model training problem into a network-enabled macro-trajectory and learning duration designs | Their deep reinforcement learning approach shows a reduction in the percentage of energy consumption compared with greedy offloading | The UAVs perform the local model training and commonality among the data across the device on the clusters only | [14] |
| Cluster-Based UAV Networks With Deep Learning (DL) | Clustering with a parameter-tuned residual network (C-PTRN) that works in two main phases, clustering and scene classification | The results show that the T2FL-C technique reaches the lowest energy consumption | The T1FL-C model method achieves poor results with a high energy consumption compared with related techniques | [15] |
| Deep Deterministic Policy Gradient (UC-DDPG) | 3D drone based on the DDPG algorithm, which considers the residual energy, mobility power, circuit power, communication power, and hover power | The simulation results show that UC-DDPG inspired by reinforcement learning has a good convergence | The RL model is not suitable for complex tasks or working in continuous and high dimensional spaces | [16] |
| Interference Management Deep Learning | Proposal of various key performance indicators (KPIs) to achieve a trade-off between maximizing the energy efficiency and spectral efficiency | The approach makes the advantages of using intelligent energy-efficient systems evident | A highly complex, non-convex optimization problem. Due to the high mobility, calculating the solution in every time instant is unrealistic | [17] |

## 3. Proposed System

In this section, we discuss our proposed system. The objective was to develop an integrated architecture that implemented a fully autonomous smart drone, able to accomplish a given mission goal in a completely independent way whilst optimizing the battery consumption to maximize the flight duration. For this purpose, we focused on a mobile architecture based on the computation at the edge paradigm, thus seeking an embedded device with acquisition and processing capabilities with appropriate characteristics. In this context, the envisaged solution leveraged on the capabilities of TinyML [18] to allow the MAV to take advantage of the ML without significantly affecting the flight duration or imposing further constraints on the already constrained MAV architecture. The block diagram of the proposed system is illustrated in Figure 1.

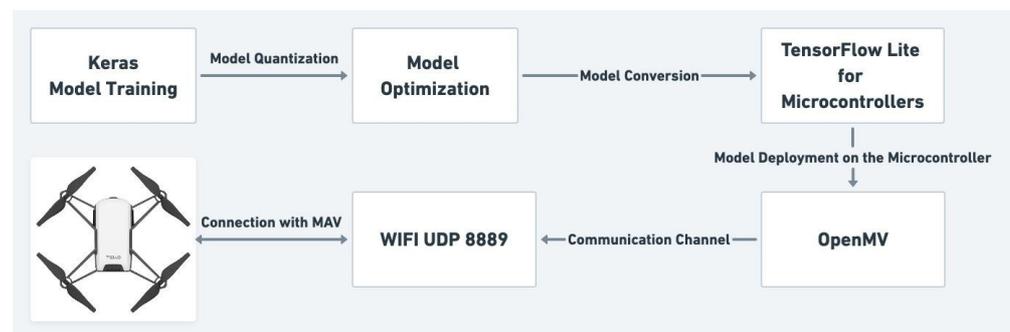

**Figure 1.** Block diagram of the proposed system.

As a basis for the development of our autonomous system, we chose a DJI Tello due to its easy programmability and wide availability. The Tello drone [19] has a maximum flight time of up to 13 minutes, a weight of about 80 g (with propellers and battery), and dimensions of 98 mm x 92.5 mm x 41 mm. It mounts 3-inch propellers and has a built-in WIFI 802.11n 2.4 G module.

As for the TinyML platform, we chose an OpenMV microcontroller, which acted as a decision unit. The OpenMV platform [20] is a small, low power microcontroller that enables the easy and intuitive implementation of image processing applications. It can be programmed using high-level Python scripts (Micro-Python). It is driven by an STM32H74VI ARM Cortex M7 processor running at 480 MHz, which is suitable for most machine vision applications. OpenMV was particularly suitable for our proposed approach due to its low power consumption and weight as OpenMV as a payload would not prevent the drone from taking off. It was also equipped with a high-performance camera that we used to collect data for the mission purposes. In addition, OpenMV was selected to ensure that the drone would be able to take off and operate whilst carrying the relevant payload due to the sharp weight limitations of the DJI Tello drone. Consequently, the energy consumption of the drone with and without the payload was tested to verify that the proposed integration was energy-effective.

The OpenMV microcontroller was connected to the drone via a WiFi UDP link to allow the data transfer. In this way, the connectivity was guaranteed independently of the availability of an internet connection, thus meeting the goal of a fully functional system regardless of the environment or situation. Communication was fundamental to control the drone through the system intelligence.

One of the most important features of the system was the TinyML model, which was integrated as part of an application to verify the functionality of the proposed system. We began building our model with the TensorFlow (TF) library, which was then converted to the lightweight version, TensorFlow Lite (TFL), which is suitable for mobile applications. The TFL model was optimized through a quantization process [21] and finally the model was deployed as part of the classification application of the microcontroller supporting the TensorFlow Lite for Microcontrollers (TFLM) framework [22]. The TFLM could interpret the model and performed the inference task based on the input data.

As for the drone navigation, the strategy was determined by the mission objective and by the results of the inference engine, which triggered a continuous detection-control loop: once an inference result was output, a decision was made on the next desired positioning of the drone and this information was communicated to the MAV to perform the necessary control operations. As an example, a detection task might require a better framing of the target object; this in turn implies the positioning of the drone in a more suitable position, which has to be translated into a set of control commands. After further target capturing, the inference engine can achieve a more accurate classification, which leads to a further target according to the overall mission goal.

After performing a few initial trials of the drone control loop, we realized that the navigation was not satisfactory due to instability problems. This in turn led to problems in capturing sufficiently stable images (e.g., motion blur, imperfect framing), often resulting in the loss of the target. To solve this problem, we introduced a PID controller in the loop [23]. The PID provided a simple but effective solution to stabilize the drone as each variable could be treated separately within a limited range where the MAV action was roughly linear. For each control iteration, we calculated the error between the actual position and the target position (which considered the position of the target to be classified). If the actual value was greater than the target value, a positive error was sent back to the system. The error was then normalized and sent to the PID, which calculated the new values for the speed of the motors. For the proportional term, we multiplied the error by a gain factor, which was set at 0.5 after several experimental trials. To control the derivative term, the difference between the current and the previous error was calculated and again multiplied by a gain factor of 0.5. Accordingly, the integral term was set by summing the errors over various steps and applying a gain factor to the cumulative error. Once we calculated all the correction factors, we transmitted the new speed values to the drone. To deal with out-of-range situations, we truncated the values in the range of +/-100.

## 4. Experimental Validation

### 4.1. Experimental Setup

In this section, we describe the experimental setup implemented to validate the proposed energy-aware smart drone in the context of a practical application. We defined a mission in which the drone navigated over a populated environment and had to identify people (face detection task) and classify them as wearing a protection mask or not (a binary classifier). The inspiration for such an application came from the current COVID-19-related recommendations as a support for local authorities to enforce relevant restrictions [24]. To implement this specific mission, we had to specialize the OpenMV microcontroller to perform the face detection and mask recognition tasks.

As for the set of data for training and validation, we adopted the open-source Face Mask Lite [25] dataset, which is based on generated faces for model training without privacy problems. In Figure 2, we represent the dataset. Even after rescaling and color conversions, image features have a high dimensionality that prevents suitable visualization. For this reason, a dimensionality reduction was applied to reduce the image to a 3D space (the three visualization layers). To train the model, we randomly selected 1000 images, equally divided into images with and without face masks. Each image was resized to 96 ×96 pixels. The number of samples was chosen to comply with the constraints of the selected microcontroller and to optimize the model size.

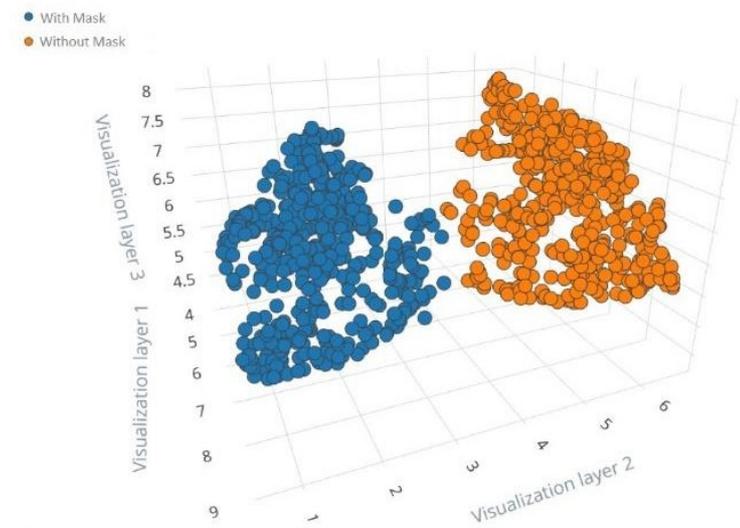

**Figure 2.** Dataset visualization and separability of the classes.

As far as the neural architecture was concerned, in our application we selected a convolutional neural network (CNN) to train a model compatible with the TinyML deployment. The face mask detection model uses MobileNet V2 architecture (see Figure 3), a well-known and proven architecture developed for image classifications [26]. MobileNet V2 consists of a stack of 16 depth-wise separable convolutional layers with an average pool followed by a fully connected layer and a SoftMax at the end. The width multiplier was chosen at 0.35 as it was optimized for our microcontroller in terms of RAM usage and an effective computation reduction.

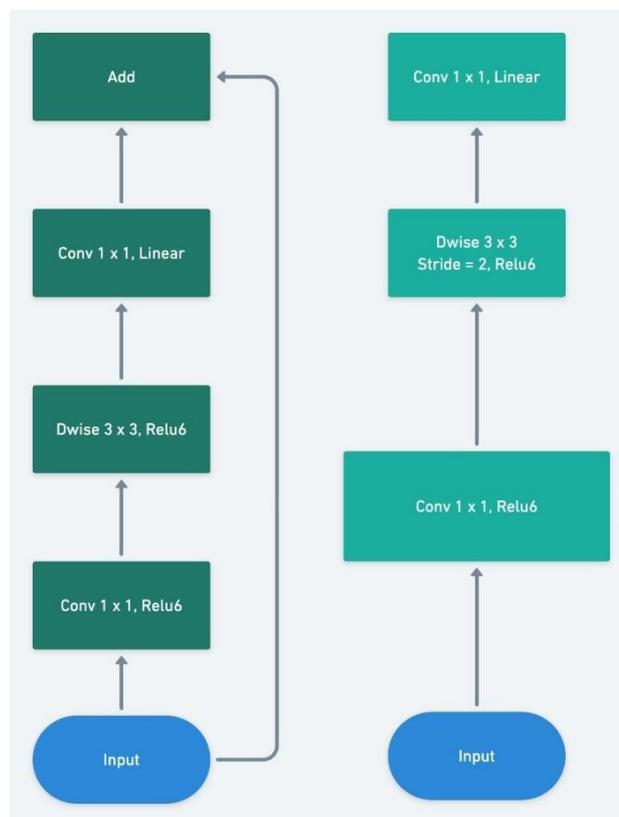

**Figure 3.** The convolutional block of MobileNet V2.

One of the critical points in bringing a deep architecture onto a tiny processor is the need to reduce as much as possible the complexity without significantly affecting the accuracy. One of the common practices consists of reducing the precision requirements for both the weights and the activations [27]. This is an essential step to compress the model to meet the hardware constraints of the OpenMV microcontroller. Although quantization is still an open research topic, it has become a standard compression method for TinyML-related applications. It allows the consumption of less flash memory and RAM whilst maintaining almost the same accuracy of the original model [28]. Moreover, compressing the network from 32-bit to 8-bit results in a significantly faster processing, shorter inference time, and lower power consumption.

In our case, the non-optimized model consumed 1.6 MB of flash memory, 1.7 s of onboard inference time, and 957 KB of RAM. The model size in this condition was not compatible with our embedded device. After introducing a dynamic quantization from a 32-bit floating point to an 8-bit integer, the resulting optimized model showed a significant reduction in size (585 KB); the onboard inference time was reduced to 859 msec and the use of RAM was limited to 296 KB with an accuracy after the post-training validation of 97%.

### 4.2. Experimental Results and Comparisons

The main parameters to assess the performance of the system concerned the performance in accomplishing the application task (classification accuracy) and the energy consumption. As for the first parameter, we should mention that the classification engine implemented in both systems was identical and the training was performed exactly in the same way. We expected, therefore, minimal differences due to numerical precision in the relevant processing architectures.

To estimate the power consumption of the drone in flight missions, we defined a detailed model of the battery performance of the drone. The measure was based on both mathematical models [see also 16] and real measures. As for the latter, we exploited the sensors mounted on the drone used for stabilization purposes and to remotely monitor the battery status of the drone (state of charge, SoC). The stability of a drone is controlled by three main sensors (i.e., gyroscopes, accelerometers, and barometers); the SoC was measured by the voltage and current sensors via a Tello software development kit (SDK).

The energy consumption of a drone varies in different states including take off, reaching the desired altitude, hovering, and maneuvering. Therefore, it was important to create a consistent test scenario for all the setups. For this purpose, all the tests were performed in equal conditions where the UAV hovered at a relative altitude of 5 m. Furthermore, each scenario was tested 10 times; the presented results were calculated by averaging the measured values.

In order to provide a comparative analysis, we selected three cases. The first set of measures referred to the drone without any payload installed. In this case, the mission was simulated with a remote controller. The second case referred to an alternative implementation that we designed using an Arduino Nano 33 BLE and the third referred to the proposed system based on the OpenMV microcontroller.

As for the Arduino implementation, it should be pointed out that this could run the same TinyML model for the same use case after going through the optimization steps using the quantization method. The Arduino Nano was chosen because of its light weight and low power consumption. The Arduino was equipped with an OV7675 camera that was used to capture data as the input to our face mask detection application. Although the Arduino was able to run the model, it took 7 seconds to perform the inference during the live classification and most of the time it failed to fully complete the classification process due to the limited RAM. Overall, the Arduino Nano did not work when the model was used for practical testing. To test the OpenMV microcontroller, we used the same model to deploy the application and run the inference. The live classification time was 859 ms with the RAM peak still below the maximum of the device. OpenMV outperformed the Arduino Nano in all comparison criteria. Nevertheless, the decision matrix for the live classification of the model showed that both microcontrollers achieved the same results. Therefore, we concluded that OpenMV was more effective for our proposed system (see Figure 4).

**Figure 4.** Processing data and confusion matrix for the Arduino Nano (above) and OpenMV (below) architectures.

Table 2 shows the average energy consumption (kJoule) and flight time (min:sec) measured across the various tests in the different setups of the system (no payload, Arduino payload, and the proposed system based on the OpenMV payload). The performances were measured in different states and flight operations.

**Table 2.** Energy consumption and flight duration comparison.

| Consumption (kJoule) | No Payload | Arduino | OpenMV |
|---|---|---|---|
| Idle | 60.192 | - | - |
| Hovering | 77.976 | 96.307 | 116.280 |
| Maneuvering | 89.727 | 112.449 | 141.588 |
| **Flying time (min:sec)** | **No Payload** | **Arduino** | **OpenMV** |
| Idle State | 12:00 | - | - |
| Hovering | 09:23 | 7:50 | 6:20 |
| Maneuvering | 08:05 | 6:42 | 5:10 |

For the idle state, the UAV was powered on without the four motors rotating; only the internal processing and LEDs were on, thus resulting in a rather stable and low power consumption. For the hovering state, the UAV was made to take off and hover at a certain altitude as hovering between altitudes caused fluctuations in the energy consumption due to sudden consumption spikes when climbing to higher altitudes. The results showed that the hovering state maintained a reasonably stable power consumption as when the UAV hovered, the forces acting on the UAV were ideally balanced. Moreover, the maneuvering state was the most energy-demanding state. In horizontal flight with a limited speed, the energy consumption fluctuated somewhat but tended to stabilize over time. The energy consumption during horizontal flight was slightly higher than the energy consumption of hovering at the same altitude. This was the result of the different thrust developed by the drone in the two different modes.

As expected, the integration of the two microcontrollers increased the energy consumption, which also affected the flight duration. Roughly speaking, the proposed setup reduced the flying time by around 30% in a typical mission involving different navigation maneuvers. Nevertheless, it should be pointed out that such a loss of performance was not dramatic compared with the advantages in terms of the capabilities of the smart system with respect to the completely manual setup.

To further validate the proposed onboard inference system, we compared it with a conventional distributed inference approach where only a camera and a communication module were installed on the drone; the processing was performed at a remote unit that communicated the results back to the UAV. In this case, the energy budget in terms of the communication link slightly decreased the performance of the distributed system. The results in terms of the energy consumption and flight time are reported in Table 3 and could be compared with the proposed system (same data of Table 2, Column 3). We noted that the distributed inference approach was slightly less energy-demanding compared with the completely onboard system and close to the Arduino-equipped setup. Nevertheless, it should be pointed out that in this case we lost the real-time capability as every decision (detection, classification, navigation) required a communication loop with a significant round-trip time. Furthermore, if the system needed to operate in an environment where it was not possible to establish a stable connection or to place a ground station, the distributed model fell short.

**Table 3.** Energy consumption and flight duration comparison with distributed processing.

| Consumption (kJoule) | Distributed Inference | OpenMV |
|---|---|---|
| Hovering | 86.320 | 116.280 |
| Maneuvering | 101.232 | 141.588 |
| **Flying time (min:sec)** | **Distributed Inference** | **OpenMV** |
| Hovering | 08:36 | 6:20 |
| Maneuvering | 07:13 | 5:10 |

## 5. Conclusions

In this paper, we presented a novel approach to endow drones with a larger autonomy and intelligence thanks to the integration of a joint flight and mission control embedded system. Thanks to the adoption of a powerful lightweight processing architecture and a suitably designed ML inference engine, the system implemented an edge computing solution that enabled the achievement of sophisticated mission goals without severely limiting the flight duration. The energy consumption of the system was tested in various setups and flying scenarios to demonstrate its energy efficiency. An experimental validation was performed on a significant sample application and showed that the impact of the integration of the microcontroller on the payload was tolerable with respect to the added value in terms of the intelligence of the system, thus making the system viable in practical contexts. Future work will include the implementation of further use cases in different application domains and the extension to other drone models to promote a broader adoption of the proposed technology.

**Author Contributions:** Conceptualization, W.R. and A.O.; methodology, W.R., A.O. and F.D.N.; software, W.R., A.O. and F.F.; validation, W.R. and A.O.; formal analysis, W.R., A.O. and F.D.N.; investigation, W.R., A.O. and F.D.N.; resources, W.R., A.O. and F.F.; data curation, W.R. and A.O.; writing—original draft preparation, W.R., A.O. and F.D.N.; writing—review and editing, W.R., A.O. and F.D.N.; visualization, W.R., A.O. and F.F.; supervision, F.D.N.; project administration, W.R and F.D.N. All authors have read and agreed to the published version of the manuscript.

**Funding:** This research received no external funding.

**Institutional Review Board Statement:** Not applicable.

**Informed Consent Statement:** Not applicable.

**Data Availability Statement:** Some or all data, models, or code that support the findings of this study are available from the corresponding author upon reasonable request.

**Conflicts of Interest:** The authors declare no conflict of interest.

## References

1. Atzori, L.; Iera, A.; Morabito, G. The internet of things: A survey. *Comput. Netw.* **2010**, *54*, 2787–2805. [CrossRef]
2. Maddikunta, P.K.R.; Hakak, S.; Alazab, M.; Bhattacharya, S.; Gadekallu, T.R.; Khan, W.Z.; Pham, Q.V. Unmanned Aerial Vehicles in Smart Agriculture: Applications, Requirements, and Challenges. *IEEE Sens. J.* **2021**, *21*, 17608–17619. [CrossRef]
3. Roberge, V.; Tarbouchi, M.; Labonté, G. Fast Genetic Algorithm Path Planner for Fixed-Wing Military UAV Using GPU. *IEEE Trans. Aerosp. Electron. Syst.* **2018**, *54*, 2105–2117. [CrossRef]
4. Qu, T.; Zang, W.; Peng, Z.; Liu, J.; Li, W.; Zhu, Y.; Zhang, B.; Wang, Y. Construction Site Monitoring using UAV Oblique Phtogrammetry and BIM Technologies. In Proceedings of the 22nd CAADRIA Conference, Suzhou, China, 5–8 April 2017.
5. Manfreda, S.; McCabe, M.F.; Miller, P.E.; Lucas, R.; Pajuelo Madrigal, V.; Mallinis, G.; Ben Dor, E.; Helman, D.; Estes, L.; Ciraolo, G.; et al. On the Use of Unmanned Aerial Systems for Environmental Monitoring. *Remote Sens.* **2018**, *10*, 641. [CrossRef]
6. Boukoberine, M.N.; Zhou, Z.; Benbouzid, M. A critical review on unmanned aerial vehicles power supply and energy management: Solutions, strategies, and prospects. *Appl. Energy* **2019**, *255*, 113823. [CrossRef]
7. Voghoei, S.; Tonekaboni, N.H.; Wallace, J.G.; Arabnia, H.R. Deep learning at the edge. In Proceedings of the 2018 International Conference on Computational Science and Computational Intelligence (CSCI), Las Vegas, NV, USA, 12–14 December 2018.
8. Banbury, C.R.; Reddi, V.J.; Lam, M.; Fu, W.; Fazel, A.; Holleman, J.; Huang, X.; Hurtado, R.; Kanter, D.; Lokhmotov, A.; et al. Benchmarking TinyML systems: Challenges and direction. *arXiv* **2020**, arXiv:2003.04821.
9. Abadi, M.; Barham, P.; Chen, J.; Chen, Z.; Davis, A.; Dean, J.; Devin, M.; Ghemawat, S.; Irving, G.; Isard, M.; et al. Tensorflow: A system for large-scale machine learning. In Proceedings of the 12th {USENIX} Symposium on Operating Systems Design and Implementation ({OSDI} 16), Savannah, GA, USA, 2–4 November 2016.
10. Ketkar, N. Introduction to keras. In *Deep Learning with Python*; Apress: Berkeley, CA, USA, 2017; pp. 97–111.
11. Marchisio, A.; Hanif, M.A.; Khalid, F.; Plastiras, G.; Kyrkou, C.; Theocharides, T.; Shafique, M. Deep Learning for Edge Computing: Current Trends, Cross-Layer Optimizations, and Open Research Challenges. In Proceedings of the 2019 IEEE Computer Society Annual Symposium on VLSI (ISVLSI), Miami, FL, USA, 15–17 July 2019; pp. 553–559. [CrossRef]
12. Lu, L.; Hu, Y.; Zhang, Y.; Jia, G.; Nie, J.; Shikh-Bahaei, M. Machine Learning for Predictive Deployment of UAVs with Rate Splitting Multiple Access. In Proceedings of the 2020 IEEE Globecom Workshops, Taipei, Taiwan, 7–11 December 2020.
13. Hoseini, S.A.; Hassan, J.; Bokani, A.; Kanhere, S.S. Trajectory optimization of flying energy sources using q-learning to recharge hotspot uavs. In Proceedings of the IEEE INFOCOM 2020-IEEE Conference on Computer Communications Workshops, Toronto, ON, Canada, 6–9 July 2020.
14. Wang, S.; Hosseinalipour, S.; Gorlatova, M.; Brinton, C.G.; Chiang, M. UAV-assisted Online Machine Learning over Multi-Tiered Networks: A Hierarchical Nested Personalized Federated Learning Approach. *arXiv* **2021**, arXiv:2106.15734.
15. Pustokhina, I.V.; Pustokhin, D.A.; Kumar Pareek, P.; Gupta, D.; Khanna, A.; Shankar, K. Energy-efficient cluster-based unmanned aerial vehicle networks with deep learning-based scene classification model. *Int. J. Commun. Syst.* **2021**, *34*, e4786. [CrossRef]
16. Qi, H.; Hu, Z.; Huang, H.; Wen, X.; Lu, Z. Energy efficient 3-D UAV control for persistent communication service and fairness: A deep reinforcement learning approach. *IEEE Access* **2020**, *8*, 53172–53184. [CrossRef]
17. Ghavimi, F.; Riku, J. Energy-efficient uav communications with interference management: Deep learning framework. In Proceedings of the 2020 IEEE Wireless Communications and Networking Conference Workshops (WCNCW), Seoul, Korea, 6–9 April 2020.
18. Sanchez-Iborra, R.; Skarmeta, A.F. Tinyml-enabled frugal smart objects: Challenges and opportunities. *IEEE Circuits Syst. Mag.* **2020**, *20*, 4–18. [CrossRef]
19. RAZE. Tello. 2019. Available online: https://www.ryzerobotics.com/kr/tello (accessed on 13 March 2021).
20. Abdelkader, I.; El-Sonbaty, Y.; El-Habrouk, M. Openmv: A Python powered, extensible machine vision camera. *arXiv* **2017**, arXiv:1711.10464.
21. Krishnamoorthi, R. Quantizing deep convolutional networks for efficient inference: A whitepaper. *arXiv* **2018**, arXiv:1806.08342.
22. David, R.; Duke, J.; Jain, A.; Reddi, V.J.; Jeffries, N.; Li, J.; Kreeger, N.; Nappier, I.; Natraj, M.; Regev, S.; et al. Tensorflow lite micro: Embedded machine learning on tinyml systems. *arXiv* **2020**, arXiv:2010.08678.
23. Joyo, M.K.; Ahmed, S.F.; Bakar, M.I.A.; Ali, A. Horizontal Motion Control of Underactuated Quadrotor Under Disturbed and Noisy Circumstances. In *Information and Communication Technology*; Springer: Singapore, 2018; pp. 63–79.
24. Mohan, P.; Aditya, J.P.; Abhay, C. Aditya, J.P.; Abhay, C. A tiny CNN architecture for medical face mask detection for resource-constrained endpoints. In *Innovations in Electrical and Electronic Engineering*; Springer: Singapore, 2021; pp. 657–670.
25. Available online: https://www.kaggle.com/prasoonkottarathil/face-mask-lite-dataset (accessed on 8 July 2021).
26. Sandler, M.; Howard, A.; Zhu, M.; Zhmoginov, A.; Chen, L.C. Mobilenetv2: Inverted residuals and linear bottlenecks. In Proceedings of the IEEE Conference on Computer Vision and Pattern Recognition (CVPR), Salt Lake City, UT, USA, 18–23 June 2018.
27. Warden, P.; Situnayake, D. *Tinyml: Machine Learning with Tensorflow Lite on Arduino and Ultra-Low-Power Microcontrollers*; O'Reilly Media: Newton, MA, USA, 2019.
28. Heim, L.; Biri, A.; Qu, Z.; Thiele, L. Measuring what Really Matters: Optimizing Neural Networks for TinyML. *arXiv* **2021**, arXiv:2104.10645.